# MOBILE ROBOTS ADAPTIVE CONTROL USING NEURAL NETWORKS

## I. Dumitrache, M. Drăgoicea

*University "Politehnica" Bucharest*
*Faculty of Control and Computers*
*Automatic Control and System Engineering Department*
*Splaiul Independentei 313, 77206 - Bucharest, Romania*
*Tel.+40 1 4119918, Fax: +40 1 4119918*
*E-mail: {idumitrache, mdragoicea}@ics.pub.ro*

Abstract: The paper proposes a feed-forward control strategy for mobile robot control that accounts for a non-linear model of the vehicle with interaction between inputs and outputs. It is possible to include specific model uncertainties in the dynamic model of the mobile robot in order to see how the control problem should be addressed taking into consideration the complete dynamic mobile robot model. By means of a neural network feed-forward controller a real non-linear mathematical model of the vehicle can be taken into consideration. The classical velocity control strategy can be extended using artificial neural networks in order to compensate for the modelling uncertainties. It is possible to develop an intelligent strategy for mobile robot control.

## 1. INTRODUCTION

Technological improvements in the design and development of the mechanics and electronics of the systems have been followed by the development of very efficient and elaborate control strategies. The framework of mobile robotics is challenging from both theoretical and experimental point of view.

The two important aspects related to mobile robotics can be addressed: a) the derivation of kinematic and/or dynamic models for wheeled mobile robots, for particular prototypes as well as for general robots equipped with wheels of several types; b) the mobile robot navigation problem that accounts for the ability of the mobile vehicle to plan and execute collision-free motions in its environment. From an engineering point of view, the problem of navigation can be addressed by taking into consideration a list of functions (collision avoidance, localisation, path-finding, representation of space, control architecture). The last function, control architecture, embodies all the other functions to allow the agent to move in its environment.

Robot autonomy requests today the leading edge of advanced robotics research. The variety of tasks to be performed by the robots from object manipulation to machine mobility, within a world environment which may be largely unknown and unpredictable impose to take into consideration improved control strategies for mobile robots. Based on the functional description of a robotic system (see figure 1), several important issues are consequently brought up such as the levels of modelling to be used, both for the task and the work environment, the computational system structure, the operating functionality and the sensor based execution control system. From an engineering point of view, the navigation problem (as a whole) of a mobile robot can be accounted for based on the following list of functions that defines elementary functionalities for navigating an autonomous mobile agent: representation of space, localisation, collision avoidance, path-finding, control architecture.

Mobile vehicles are non-holonomic systems for which the control problem was addressed in (Campion, *et al.*, 1991), (Bloch and Reyhanoglu, 1992), (d'Andrea Novel, *et al.*, 1995). Many other theoretical control problems have been studied concerning mobile vehicles, the difficulty of the control problem depending also on the control objective, not only on the non-holonomic nature of the system. Different architectures of autonomous control systems for mobile robots taking into consideration intelligent techniques like fuzzy logic, neural networks, genetic algorithms and symbolic AI are presented in the literature (Tunstel *et al.*, 1998).

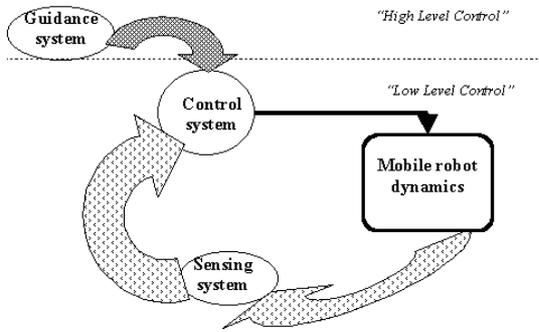

Figure 1 Functional description of a robotic system

Even though significant progress has been made in the study of such non-holonomic mechanical systems, both in the topic of modelling and control, there are still some problems to be addressed. One of this refers to the control problem of non-holonomic systems when the model of the system, in particular, of the mobile vehicle, is affected by uncertainties, arising from neglected dynamics or parameter variations. In such cases, the performances of any control strategy that uses such model will be affected. This article proposes a control strategy based on neural network architectures, that can take into consideration this modelling uncertainties. Researches made in the recent years demonstrate that artificial neural networks (ANN) offer a great promise for the modelling and control of non-linear systems. Even though the required network complexity for non-linear control applications and the computational expenses are still important problems to be solved, more and more results are reported in the field of neural control, such that ANN proved to be a real challenge for modern engineering technology that have to deal with increasingly complex space structures with ever more demanding performance criteria.

## 2. THE PROCESS

A car-like mobile robot rolling on a flat ground was selected as a case study for the proposed control strategy.

This type of vehicle can be represented by a rectangular shape moving in $R^2$ (figure 2). The movement of the robot in an empty space can be described by the posture vector $\xi = (x, y, \theta)$, where $(x, y)$ are the coordinates of the midpoint P between the two rear wheels, and $\theta \in [0, 2\pi)$ is the rotation angle between the coordinate system that does not move with the robot $(x_0\, O_0\, y_0)$ and the coordinate system attached to the robot $(x_r P y_r)$.

The linear velocity along the main axis of the robot ($V_{AV}$) and its rotation velocity ($\dot{\theta}$) are the two parameters that determines the instantaneous motion of this type of vehicle. Based on these, the kinematic state model of the robot can be defined (Betourne, 1995):

$$\begin{aligned}
\dot{x}_1 &= \cos(x_3)\cos(x_4)u_1 \\
\dot{x}_2 &= -\sin(x_3)\cos(x_4)u_1 \\
\dot{x}_3 &= \frac{1}{L}\sin(x_4)u_1 \\
\dot{x}_4 &= u_2 \\
\dot{x}_5 &= \frac{1}{r}u_1 \\
\dot{x}_6 &= \frac{1}{r}\bigl(\cos(x_4) + (D/L)\sin(x_4)\bigr)u_1 \\
\dot{x}_7 &= \frac{1}{r}\bigl(\cos(x_4) - (D/L)\sin(x_4)\bigr)u_1
\end{aligned} \qquad (1)$$

where the state vector is $\underline{x} = [x\; y\; \theta\; \phi\; \varphi_c\; \varphi_s\; \varphi_d] \in R^7$. With the above mentioned definitions, the equations of the steering system can be rewritten:

$$\begin{bmatrix}\dot{x}\\ \dot{y}\\ \dot{\theta}\end{bmatrix} = \begin{bmatrix}\cos\theta & 0\\ \sin\theta & 0\\ 0 & 1\end{bmatrix}\begin{bmatrix}\eta_1\\ \eta_2\end{bmatrix} = S(\underline{q})\underline{\eta} \qquad (2)$$

where $\underline{\eta} = \begin{bmatrix} \eta_1 \\ \eta_2 \end{bmatrix} = \begin{bmatrix} V_{AV} \\ \dot{\theta} \end{bmatrix}$, $|\eta_1| \leq V_{Av\_max}$ and $|\eta_2| \leq \dot{\theta}_{max}$.

The entries of the vector $\underline{\eta}$ represent the velocity control inputs, and $x_4 = \phi$ is an internal state associated to the orientable wheel of the robot.

## 3. NEURAL NETWORKS BASED CONTROL STRATEGY

The most important function necessary for navigating an autonomous agent, the control architecture, embodies all the other mentioned functions and makes the agent to be "alive". In particular, this work addresses a trajectory tracking problem (Sarkar, 1994) in the presence of modelling uncertainties of the controlled system. An important control problem for mobile robots is to include certain specific model uncertainties in the dynamic model of the mobile robot. It seems attractive to initiate an adaptive control for parameter uncertainties (e.g., unknown mass, wheel-radius; etc…).
The aim of the present paper is to formulate a control strategy for mobile robot control based on artificial neural networks (ANN). The approach considered in the present paper combines the stabilisation properties of an feedback conventional PID controller with the above mentioned learning capabilities of a neural network based feed-forward controller.

Conceptually, most of the approaches that accounts for neural networks based control structures make use of the "inverse" model of the process as a controller. From this category, the two strategies that gained increased importance lately are IMC (Internal Model Control) and feed-forward control. In this article the feed-forward controller is implemented as a neural network controller that represents the inverse model of the process.

The mobile robot can be considered to be a two inputs - (the torques for the dc. motors) - two outputs (the linear velocity along the main axis, $V_{AV}$, and the rotation velocity, $\dot{\theta}$) system.

The two levels hierarhical control structure is presented in figure 3. The inner loop is used to control the velocities of the wheels (velocity control), while the outer loop forces the vehicle to move on the desired trajectory (motion control). The motion controller is the generator of the velocities references for the velocity controller.

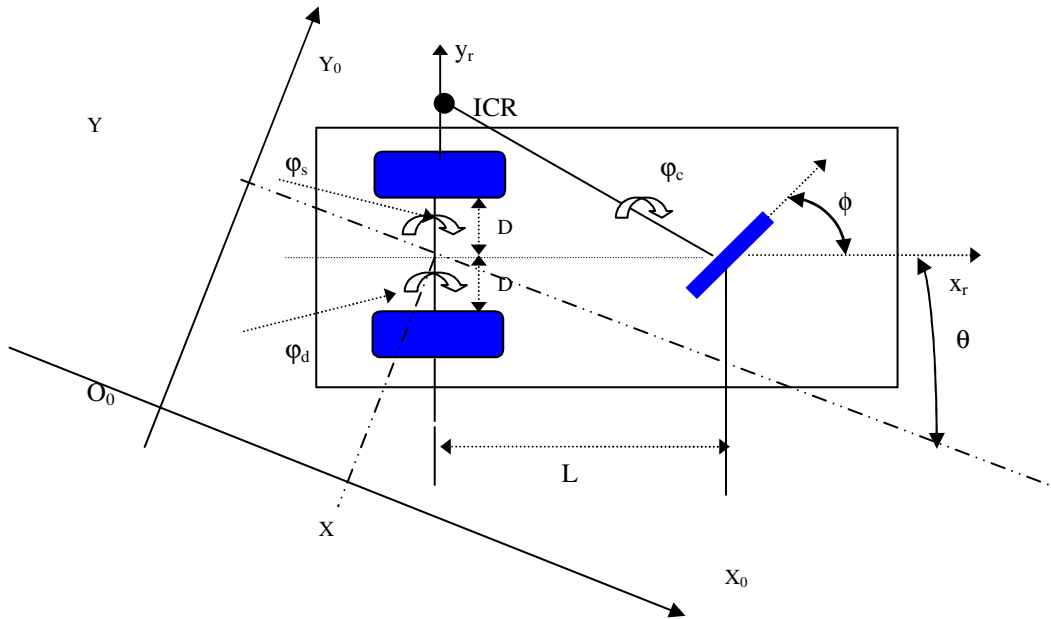

Figure 2 Posture definition

A feed-forward adaptive control strategy based on neural networks is used in order to implement the velocity controller. Figure 3 depicts the two components of the velocity controller:

- a feedback component (PID control of linear and angular velocities, $V_{AV}$ and $\dot{\theta}$), based on a

simplified model of the process (dc motors and mobile platform) (equation 3) (Mazo, 1995); it will stabilise the process and will reject the perturbations.

$$M\ddot{V}_{AV} = \frac{1}{2r}(P_d + P_s)$$
$$J\ddot{\theta} = \frac{r}{2D}(P_d - P_s) \quad (3)$$
$$P_{s,d} = N\left(\frac{k_M}{R_a}U_{s,d} - \frac{k_M k_E}{R_a}\frac{60N}{2\pi r}V_{s,d} - k_M I_0\right)$$

where $P_{l,r}$ are the left/right torques obtained from the dc motors, M is the mass of the platform, r radius of the wheels, D half distance between rear wheals, J the inertia momentum of the platform, N gear ratio, $k_M$ torque constant, $R_a$ terminal rezistance, $k_E$ back emf constant, $I_0$ no load current, $V_{l,r}$ left/right wheels velocities.

- a feed-forward component, with learning abilities, will allow a fast trajectory tracking. This component is important in improving the system performances by learning on-line information about the process, through direct interaction. Actually, this component will learn on-line the inverse dynamics of the process.

In this way, it is possible to implement an adaptive control strategy at the execution level, based on neural networks. The neural network (feed-forward controller) generates a command signal $U_{ff}$ that will adjust the signal generated by the feedback controller, $U_{fb}$, in order to minimise the velocity reference error, while compensating the modelling uncertainties. The original contribution of this work consists in obtaining the command vector $\underline{u} = [U_l \; U_r]$, using the reference velocity vector $\underline{\eta}_c$ from the motion controller. The motion controller was implemented following the procedure presented in (Kanayama, 1990).

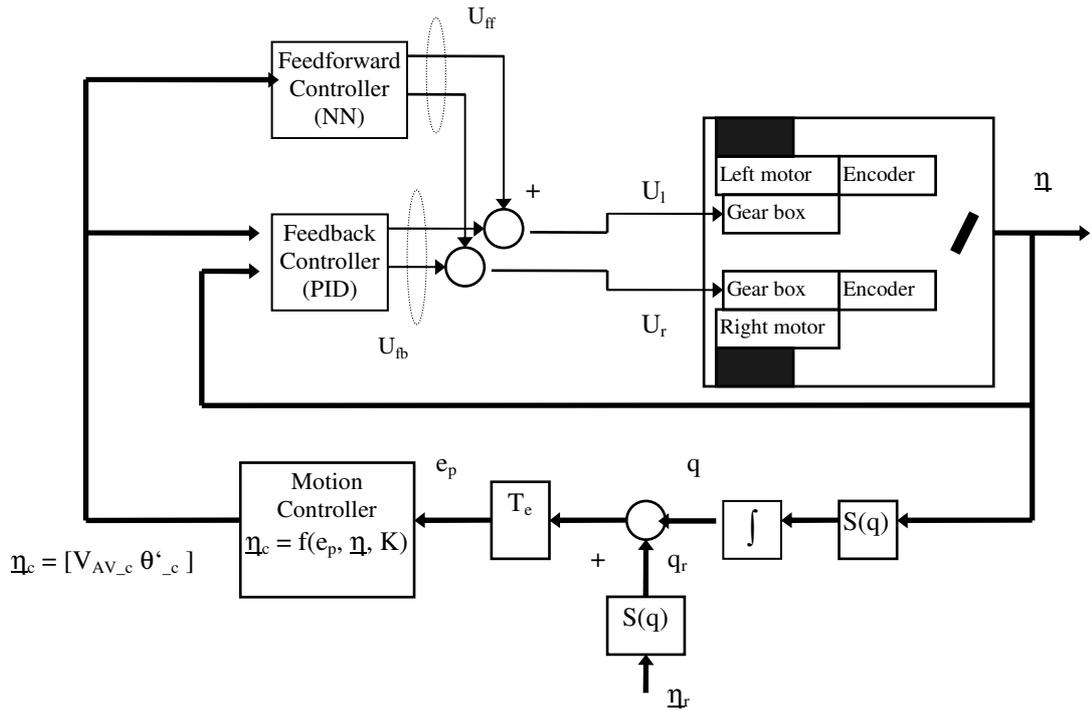

Figure 3 Two level hierarchical control structure

Some results are depicted in figures 4 and 5. Figure 4 presents results obtained using a feed-forward control strategy (average velocity, rotation velocity and tracking errors). Figure 5 presents, for comparison, the tracking errors for the two strategies (conventional PID and feed-forward with neural networks).

The typical evolution of the neural network outputs (the feed-forward control signal $U_{ff}$) is presented in figure 6.

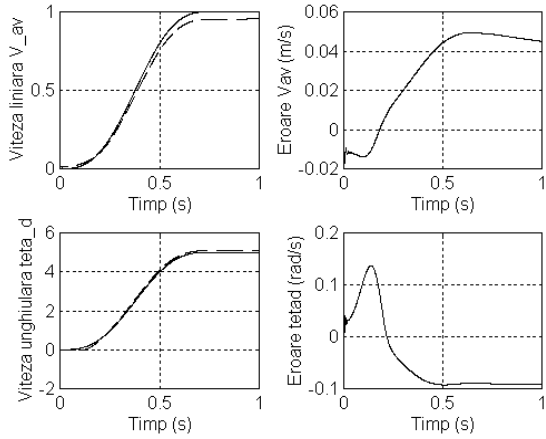

Figure 4 a)

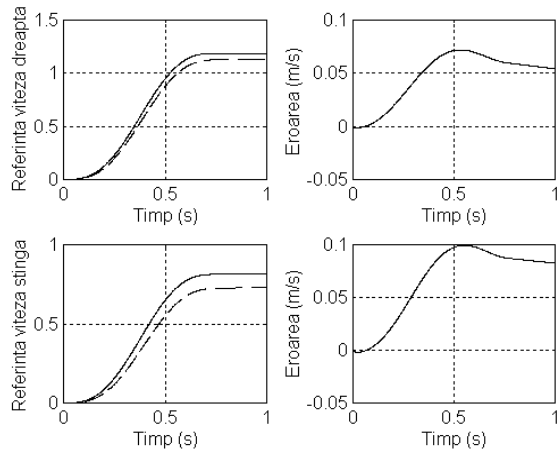

Figure 4 b)

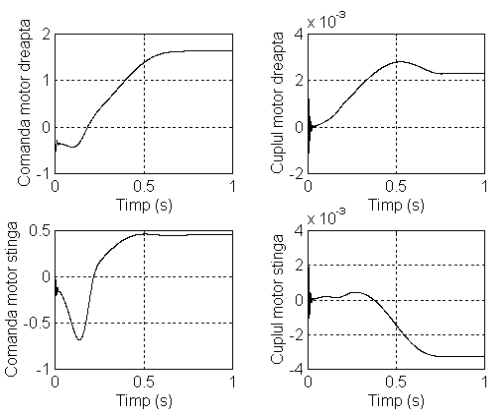

Figure 4 c)Velocity control - neural networks feed-forward control strategy

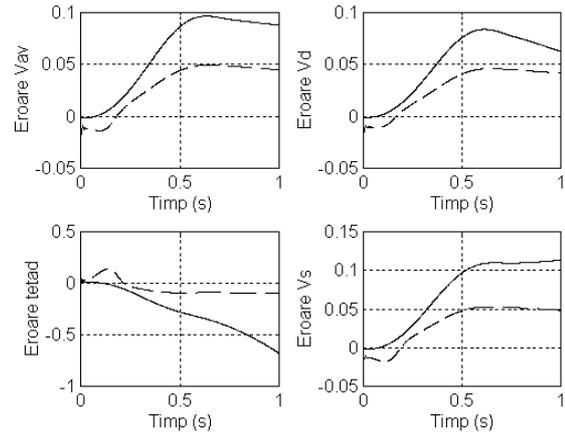

Figure 5 Tracking errors - conventional PID control versus neural networks feed-forward control

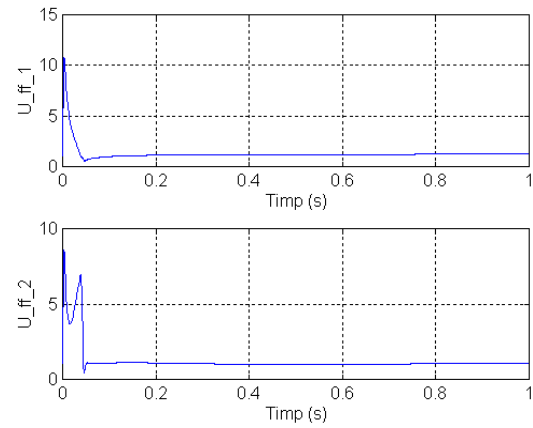

Figure 6 Neural network feed-forward controller outputs

## 4. CONCLUSIONS

This paper proposes a feed-forward control strategy for mobile robot control that accounts for the real non-linear model of the vehicle with interaction between inputs and outputs. In this respect, it is possible to include specific model uncertainties in the dynamic model of the mobile robot in order to see how the control problem should be addressed taking into consideration the complete dynamic mobile robot model. By means of a neural network feed-forward controller a real non-linear mathematical model of the vehicle can be taken into consideration. The classical velocity control strategy was extended using artificial neural networks in order to compensate for the modelling uncertainties. In this way it is possible to develop an intelligent strategy for mobile robot control.

.